\pdfoutput=1

\documentclass[11pt]{article}

\usepackage{acl}

\usepackage{times}
\usepackage{latexsym}

\usepackage[T1]{fontenc}

\usepackage[utf8]{inputenc}

\usepackage{microtype}

\usepackage{inconsolata}

\usepackage{boxedminipage} 
\usepackage{xcolor} 
\usepackage{graphicx}
\usepackage{multirow}
\usepackage{multicol}

\usepackage{array}
\usepackage{arydshln}

\usepackage{booktabs}

\definecolor{gred}{RGB}{219,68,55}
\definecolor{gblue}{RGB}{66,133,244}
\definecolor{gyellow}{RGB}{244,180,0}
\definecolor{ggreen}{RGB}{15,157,88}
\definecolor{ggrey}{RGB}{115,115,115}

\usepackage{soul}   

\usepackage{caption}
\usepackage{subcaption}
\usepackage{tabularx}

\usepackage{amsmath}

\usepackage[normalem]{ulem} 

\usepackage{adjustbox}

\let\svthefootnote\thefootnote
\newcommand\freefootnote[1]{%
  \let\thefootnote\relax%
  \footnotetext{#1}%
  \let\thefootnote\svthefootnote%
}

\usepackage{bbding}

\usepackage{svg}
\svgpath{{./figures/}} 

\usepackage{listings}
\usepackage{booktabs}
\definecolor{codegreen}{rgb}{0,0.6,0}
\definecolor{codegray}{rgb}{0.5,0.5,0.5}
\definecolor{codepurple}{rgb}{0.58,0,0.82}
\definecolor{backcolour}{rgb}{0.95,0.95,0.92}

\lstdefinestyle{mystyle}{
    xleftmargin=3mm,
    commentstyle=\color{codegreen},
    keywordstyle=\color{magenta},
    numberstyle=\tiny\color{codegray},
    stringstyle=\color{codepurple},
    basicstyle=\ttfamily\footnotesize,
    breakatwhitespace=false,         
    breaklines=true,                 
    captionpos=b,                    
    keepspaces=true,                 
    numbers=left,                    
    numbersep=5pt,                  
    showspaces=false,                
    showstringspaces=false,
    showtabs=false,                  
    tabsize=2
}

\usepackage{makecell}

\usepackage{fancyvrb}

\usepackage{todonotes}

\usepackage{subcaption,ragged2e}

\title{Neuro-Symbolic Integration Brings Causal and Reliable Reasoning Proofs~\thanks{\hspace{5pt}The work described in this paper is partially supported by a grant from the Research Grant Council of the Hong Kong Special Administrative Region, China (Project Code: 14200620).} ~\thanks{\hspace{5pt}This work was supported by Alibaba Group through the Alibaba Innovative Research (AIR) Program (TA2217728).}
}

\author{
Sen Yang~$^{1~2}$ \hspace{14pt} Xin Li~$^{2}$~\thanks{\hspace{5pt}XL is the corresponding author. } \hspace{14pt} Leyang Cui \hspace{14pt} Lidong Bing~$^{3}$ \hspace{14pt} Wai Lam~$^1$ \\
$^1$~The Chinese University of Hong Kong \\
$^2$~DAMO Academy, Alibaba Group \\
$^3$~Shanda AI Research Institute \\
\texttt{senyang.stu@gmail.com} \hspace{8pt}
\texttt{xinting.lx@alibaba-inc.com} \\
\texttt{lidong.bing@shanda.com} \hspace{18pt}
\texttt{neal19951101@gmail.com} \\
\texttt{wlam@se.cuhk.edu.hk}
}

\begin{document}
\maketitle
\begin{abstract}
Two lines of approaches are adopted for complex reasoning with LLMs.
One line of work prompts LLMs with various reasoning structures, while the structural outputs can be naturally regarded as intermediate reasoning steps.
Another line of work adopt LLM-free declarative solvers to do the reasoning task, rendering higher reasoning accuracy but lacking interpretability due to the black-box nature of the solvers.
Aiming to resolve the trade-off between answer accuracy and interpretability, we present a simple extension to the latter line of work.
Specifically, we showcase that the intermediate search logs generated by Prolog interpreters can be accessed and interpreted into human-readable reasoning proofs.
As long as LLMs correctly translate problem descriptions into Prolog representations, the corresponding reasoning proofs are ensured to be causal and reliable.
On two logical reasoning and one arithmetic reasoning datasets, our framework obtains significant improvements in terms of both answer accuracy and reasoning proof accuracy. 
Our code is released at \url{https://github.com/DAMO-NLP-SG/CaRing}.
\end{abstract}

\section{Introduction}
\label{sec:intro}

\begin{figure*}[t!]
    \centering
    \includegraphics[width=0.95\textwidth]{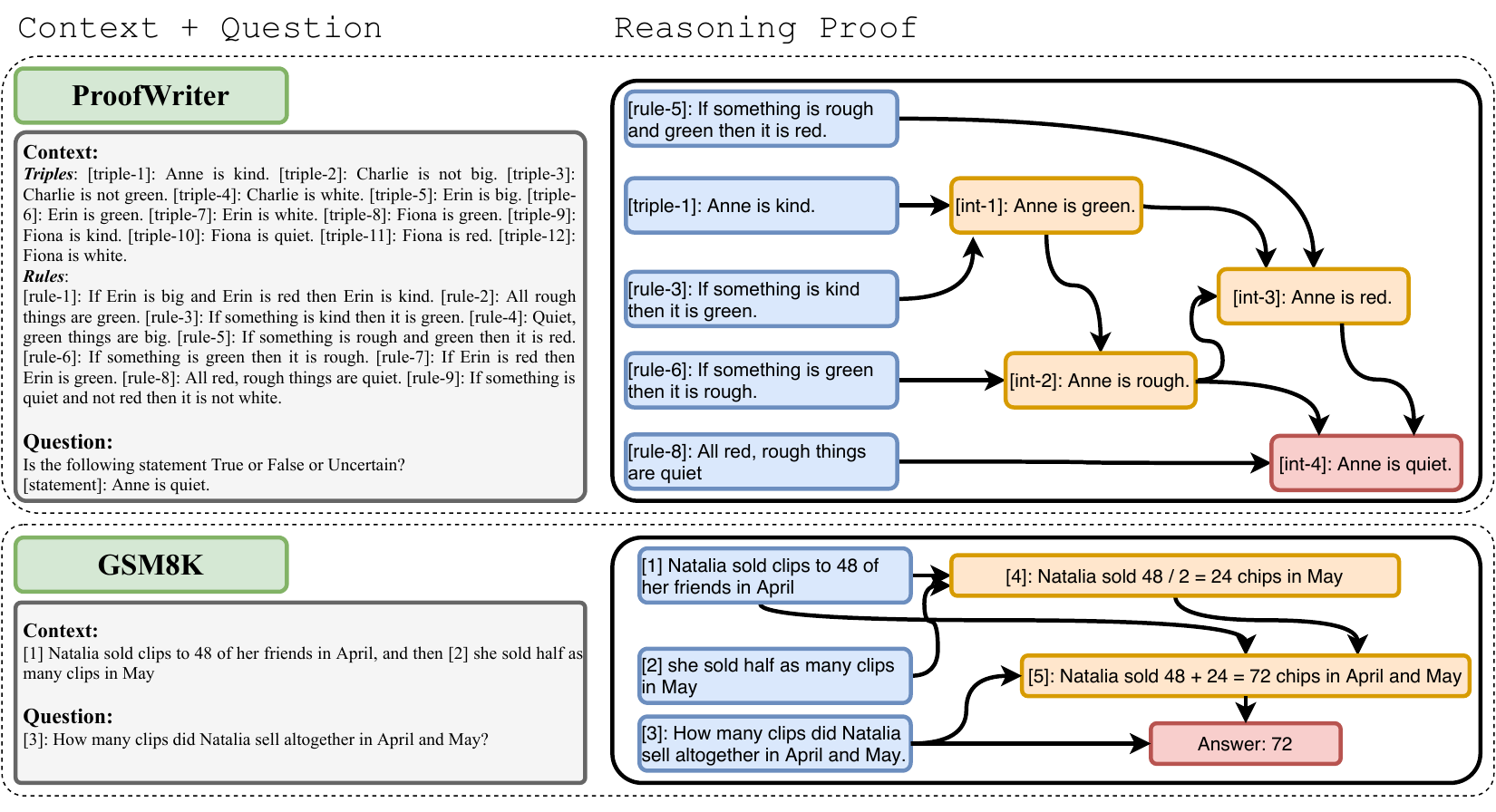}
    \caption{
    Two examples of complex/structured reasoning problems from ProofWriter and GSM8K, respectively.
    The reasoning proofs in such problems formulate directed acyclic graphs (DAGs) in a multi-step and multi-premise manner.
    }
    \label{fig:method:problem}
    \vspace{-6pt}
\end{figure*}

Large language models (LLMs), like LLaMA-3~\citep{dubey2024llama3herdmodels} and GPT-4~\citep{gpt4}, are shown to be powerful, but they still struggle with structurally complex reasoning problems, such as logical reasoning~\citep{tafjord-etal-2021-proofwriter} and complex arithmetic reasoning~\citep{cobbe2021gsm8k,ribeiro2023street}, as shown in Figure~\ref{fig:method:problem}.
Related work argued that such defects result from the fact that existing LLMs fall short on planning~\citep{valmeekam2023planbench,huang2024large,kambhampati2024llmscantplanhelp}.
This is predictable, since one should not expect a next-token-prediction LLM to solve an $n$-step reasoning problem whose reasoning space grows exponentially regarding $n$.

Existing LLM-based approaches targeting at this issue can be divided into two-fold. 
One line of work, such as Self-Consistency~\citep{selfconsistency}, Tree-of-Thoughts~\citep{yao2023treeofthoughts} and RAP~\citep{rap}, adopt various search strategies (e.g., Monte Carlo tree search) to iteratively prompt LLMs, expecting that the correct reasoning path could be achieved using proper process supervision that guides the searching process.
Their intermediate search steps are explicit, making the reasoning process interpretable.
However, they still rely heavily on LLMs themselves to do planning, thus not fully addressing the aforementioned issue.
Besides, LLM inference are performed in each searching step, while solving one problem usually requires many search steps, leading to excessive computational cost. 
Another line of work, such as Logic-LM~\citep{PanLogicLM23} and SatLM~\citep{ye2023satlm}, translate problem descriptions into declarative symbolic representations (i.e., code) and then adopt off-the-shelf solvers to execute the symbolic representations.
Such methods enjoy two key advantages: (1) The declarative symbolic representation is closer to the problem description than the reasoning steps are, so it would be easier for LLMs to do problem translation rather than producing the reasoning steps themselves. (2) The reasoning task is undertaken by an external solver, guaranteeing the correctness of the reasoning process as long as the symbolic representations are correct.
However, their search steps are implicitly performed within the solver, making it difficult for humans to inspect whether the reasoning process is correct.

Tracking the above, this work presents a framework that gets rid of LLMs when performing the reasoning task while still yielding interpretable reasoning steps.
Our idea is simple: Searching with symbolic solvers record automatically-generated logs (i.e., trace), which can be translated into human-readable intermediate steps upon further processing.
Following this idea, we implement an approach using Prolog, a declarative programming language based on formal logic.
We evaluate our approach, \textsc{CaRing} (\underline{Ca}usal and \underline{R}eliable Reason\underline{ing}), on three reasoning datasets that contain reasoning proof annotations, including two logical reasoning datasets, ProofWriter~\citep{tafjord-etal-2021-proofwriter} and PrOntoQA~\citep{saparov2023language}, and one arithmetic reasoning dataset, GSM8K~\citep{ribeiro2023street}.
\textsc{CaRing} consistently outperforms the CoT baseline and existing methods in terms of answer accuracy, reasoning proof similarity, and reasoning proof accuracy.
On the challenging ProofWriter dataset, \textsc{CaRing} using Code-LLaMA-34B yields an answer accuracy of 96.5\% and a reasoning proof similarity of 81.0\%, while previous SoTA achieved an answer accuracy of 79.7\%.
Further analysis indicates \textsc{CaRing} remains robust when the reasoning problem becomes more complex.

\section{Related Work}



\subsection{Explainable Complex Reasoning}

The Chain-of-Thought prompting method, which found out reasoning with LLMs benefits from generating intermediate steps, has sparked a recent trend in how to better do reasoning while remaining explainable.
Several works investigated using other reasoning structures, such as trees~\citep{yao2023treeofthoughts,long2023tot} and graphs~\citep{graphofthoughts,zhang2023cumulative}.
These approaches have shown improved performance, particularly in complex reasoning tasks where the processes involved are often more intricate than simple linear chains. However, despite the alignment of their reasoning proof structures with the gold-standard proofs, these methods still face challenges in ensuring causality and reliability. This limitation stems from their reliance on LLMs for deliberate reasoning, which are prone to hallucinations and may compromise causality.

Some other recent works adopted a less structured manner~\citep{sanyal-etal-2022-fairr,tafjord-etal-2022-entailer,creswell2023selectioninference,kazemi-etal-2023-lambada}.
For example, Selection-Inference~\citep{creswell2023selectioninference} divides the reasoning process into two phases: (1) the Selection phase for selecting the premises that might be relevant for the next round of inference, and (2) the Inference phase for conducting a single reasoning step with the selected knowledge fragments.

\subsection{Neuro-symbolic Reasoning}
Neuro-symbolic systems attempt to leverage the strengths of both neural networks and symbolic reasoning~\citep{NeuralModularNetworks,neelakantan2017learning,hudson2019learning,Gupta2020NeuralModular,nye2021improving}.
This includes the use of neural networks for pattern recognition and learning from unstructured data, integrated with symbolic systems for rule-based reasoning and knowledge representation.
Despite significant progress, neuro-symbolic reasoning faces challenges, notably in scalability and the efficient integration of learning and reasoning components. 

Recent advancements in neuro-symbolic research, particularly in reasoning over text, have utilized LLMs to encapsulate knowledge from unstructured human languages, as noted in \citet{lyu2023faithful,PanLogicLM23}.
These methods typically translate natural language into symbolic representations for subsequent execution-based reasoning.
However, they have not fully explored the capabilities of symbolic solvers in generating detailed reasoning proofs.
In contrast, our approach leverages customized meta-interpreters in conjunction with symbolic solvers to uncover and articulate the underlying reasoning proofs.
This not only enhances the transparency of automatic reasoning systems but also simplifies the process for humans to verify their correctness and safety.

\section{Method}
\label{sec:method}

The problem we are interested in is featured with structured or complex reasoning.
As depicted in Figure~\ref{fig:method:problem}, these problems typically necessitate multi-step and multi-premise reasoning over a directed acyclic graph (DAG), where individual nodes signify distinct knowledge fragments and directed edges denote reasoning steps.
Each reasoning step uses existing knowledge to infer new relevant knowledge.
Numerous knowledge fragments are often aggregated to infer a new one, which we denote as ``multi-premise''.
The solver usually performs multiple such steps to reach an ultimate goal, which we denote as ``multi-step''.
This entire reasoning process naturally composes a DAG.

We are interested in providing accurate answers along with causal and reliable explanations for such reasoning problems.
\citet{algorithm_logic_control} proposed that \textit{Algorithm = Logic + Control}, where \textit{Logic} refers to the knowledge which can be used to solve the problem and \textit{Control} refers to the problem-solving strategy in which the knowledge can be used.
They further proved that an algorithm benefits from separating the \textit{Logic} component and the \textit{Control} component.
Existing works that adopt a symbolic solver to search for valid reasoning paths have followed this design principle~\citep{PanLogicLM23,ye2023satlm}.
Our work also follows this principle.

\begin{figure}[t!]
    \centering
    \includegraphics[width=0.95\columnwidth]{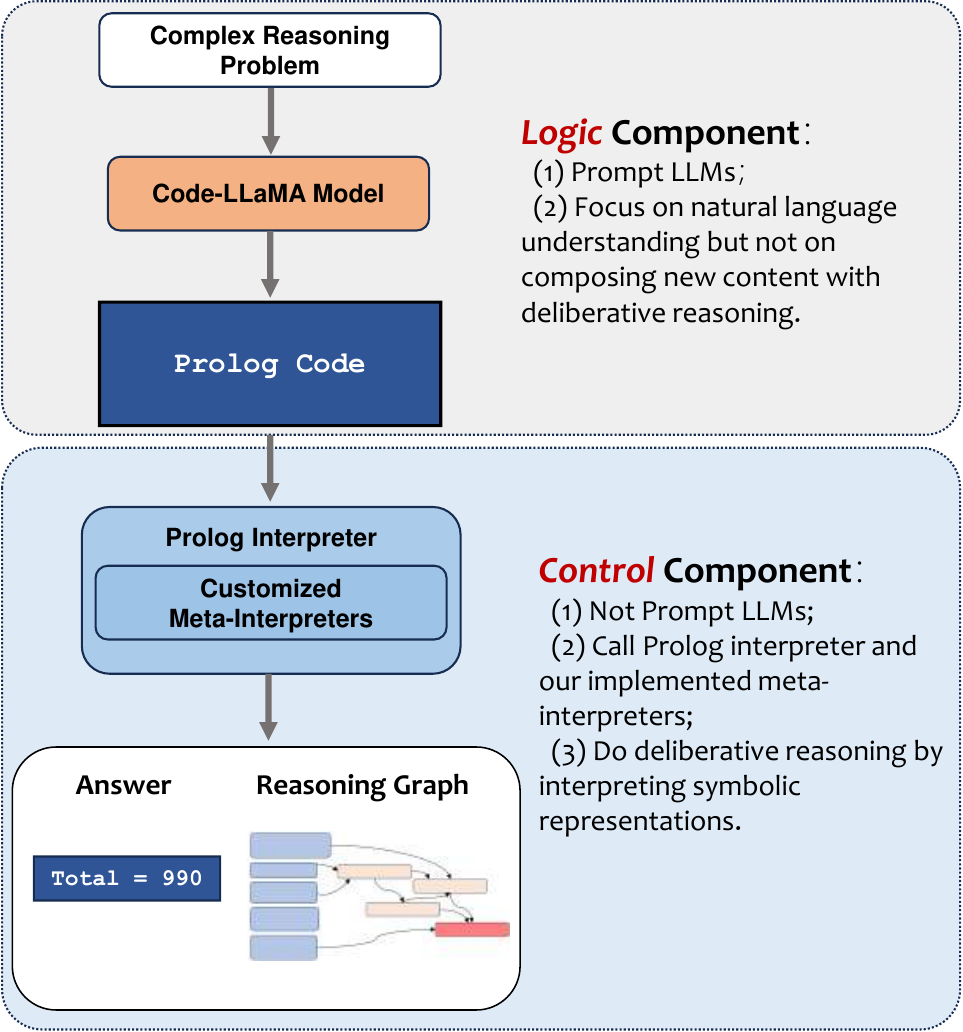}
    \caption{
    Illustration of our \textsc{CaRing} framework, consisting of a \textit{Logic} component and a \textit{Control} component.
    }
    \label{fig:method:method}
\end{figure}

We present \textsc{CaRing} (\underline{Ca}usal and \underline{R}eliable Reason\underline{ing}), a modular approach consisting of two components: 
\begin{itemize}
    \item \textsc{SymGen}: LLM-based symbolic representation generator ($\S$\ref{sec:method:code_translate}), which translates problem descriptions into formal symbolic knowledge representations that can be used for symbolic inference. 
    \item \textsc{SymInfer}: LLM-free symbolic inference engine ($\S$\ref{sec:method:symbolic_inference}), which performs deliberate reasoning by executing the symbolic representations provided by \textsc{SymGen}. 
    With customized Prolog meta-interpreters, \textsc{SymInfer} supports 
    (i) causal and reliable tracing of the reasoning process ($\S$\ref{sec:method:symbolic_inference:reason_trace}); 
    (ii) various search strategies, such as Depth-First Search (DFS) and Iterative Deepening Search (IDS) ($\S$\ref{sec:method:symbolic_inference:search_strategy}).
    
\end{itemize}

\begin{figure}[t!]
    \centering
    \includegraphics[width=0.8\columnwidth]{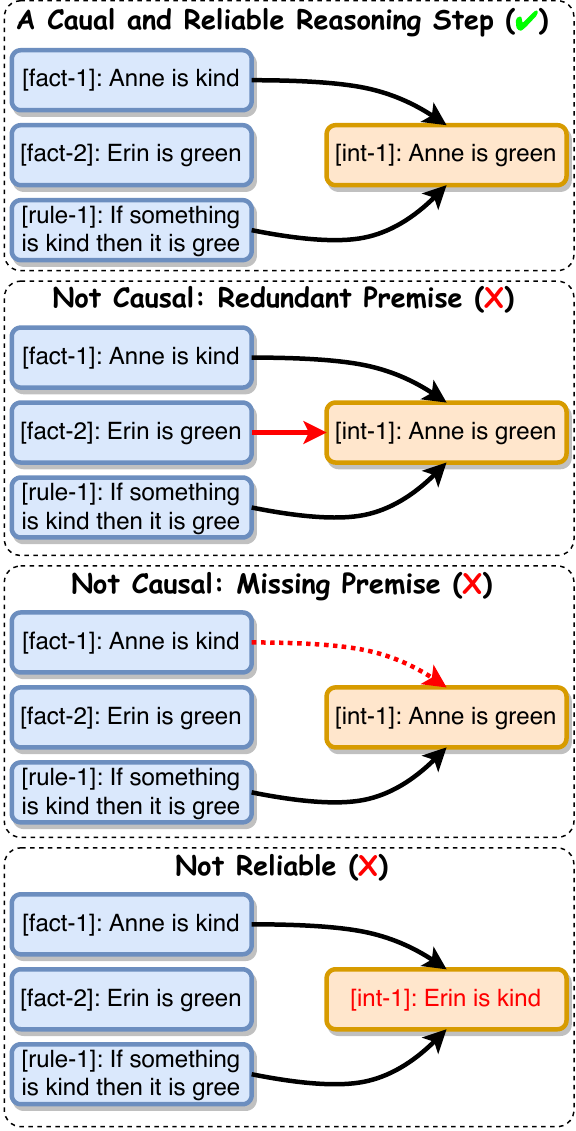}
    \caption{
    Illustrations of how causality and reliability play important roles in reasoning.
    LLMs may be (i) non-causal by selecting redundant premises or ignoring relevant ones and (ii) non-reliable by hallucinating erroneous contents.
    }
    \label{fig:intro:causality_and_reliability}
\end{figure}

The machine-execution nature of \textsc{SymInfer} guarantees both causality and reliability, as shown in Figure~\ref{fig:intro:causality_and_reliability}.
Under the principle of \textbf{Causality}, the inference of a new knowledge piece is strictly linked to those existing fragments that are relevant, ensuring precise and limited attribution.
This implies that a causal relationship is established only when the preceding event (at the base of the edge) directly influences the subsequent event (at the apex).
Regarding \textbf{Reliability}, the content within each newly inferred node is the result of a deterministic process, safeguarding it from the kinds of erroneous hallucinations often encountered in outputs from LLMs.

\begin{table}[t]
\centering
\small
\begin{tabular}{ll}
\specialrule{.12em}{1em}{0em}
    \textbf{Natural Language} & \textbf{Prolog Code} \\
    \hline
    \textit{Fiona is green. } & 
    \begin{lstlisting}
green(fiona). \end{lstlisting} \\
    \cdashline{1-2}[1.2pt/2pt]
    \makecell[l]{\textit{All red, rough things}\\ \textit{are quiet. }} & 
    \begin{lstlisting}
quiet(X) :-
    red(X), rough(X). \end{lstlisting} \\
    \cdashline{1-2}[1.2pt/2pt]
    \makecell[l]{\textit{Tina makes \$18.00}\\ \textit{an hour. }} &
    \begin{lstlisting}
wage(18.00). \end{lstlisting} \\
    \cdashline{1-2}[1.2pt/2pt]
    \makecell[l]{\textit{( she is eligible for}\\ \textit{overtime,) which is}\\ \textit{paid by your hourly}\\ \textit{wage + 1/2 your}\\ \textit{hourly wage. }} &
    \begin{lstlisting}
overtime_wage(W) :-
    wage(W1),
    W is 1.5 * W1. \end{lstlisting} \\
\specialrule{.12em}{.05em}{.0em}
\end{tabular}
    \caption{Examples of problem description snippets and their Prolog representations. It can be seen that the Prolog code is highly declarative, rendering less challenges for LLMs. In other words, the LLM does not need to infer new knowledge, thus avoiding hallucination or planning reasoning paths. }
    \label{tab:method:example_code}
\end{table}

\subsection{\textsc{SymGen}: Symbolic Representation Generator}
\label{sec:method:code_translate}

To represent \textit{Logic} (i.e., the knowledge which can be used to solve the problem), we adopt a popular logic programming language, Prolog~\citep{prolog}.
Prolog is a declarative programming language, in which \textit{Logic} is expressed as relations (called Facts and Rules), with several examples shown in Table~\ref{tab:method:example_code}.
A computation is initiated by running a query over these relations, which will be further explained in $\S$\ref{sec:method:symbolic_inference}.

Though LLMs are prone to hallucinate erroneous facts when composing new knowledge (i.e., generating reasoning steps), they are shown to be powerful at understanding natural languages and directly translating them into other formats~\citep{ye2022the,saparov2023language}.
Prompting LLMs to translate problem descriptions into symbolic representations just enjoys this strong point of LLMs, as previously shown by \citet{ye2023satlm}.
As for implementation, we few-shot prompt LLMs with several human-written in-context demonstrations, each containing a problem and corresponding Prolog representations.

\subsection{\textsc{SymInfer}: Symbolic Inference Engine}
\label{sec:method:symbolic_inference}

We use \textsc{SymInfer} to produce answers and reasoning traces by executing the aforementioned symbolic representations.
Since we adopt Prolog to represent knowledge, our symbolic inference engine is naturally instantiated with Prolog interpreters.
By default, the SWI-Prolog~\citep{swipl} interpreter adopts the Depth-First Search (DFS) backtracking strategy and does not yield reasoning proofs.
We adopt customized Prolog-based meta-interpreters\footnote{The meta-interpreters used in this work are built upon the implementations of \citet{Triska}. } to achieve two goals: 
(i) To produce reasoning proofs;
(ii) To adopt better search algorithms other than DFS.

\subsubsection{Reasoning Tracer}
\label{sec:method:symbolic_inference:reason_trace}
Below is a Prolog meta-interpreter to show the reasoning proofs~\citep{Triska}:

\begin{lstlisting}[language=Prolog]
% Define the operator for proofs
:- op(750, xfy, =>).

% Proof tree generation
mi_tree(true, true).
mi_tree((A,B), (TA,TB)) :-
        mi_tree(A, TA),
        mi_tree(B, TB).
mi_tree(G, builtin(G)) :-
        predicate_property(G, built-in),
        !, 
        call(G).
mi_tree(g(G), TBody => G) :-
        mi_clause(G, Body),
        mi_tree(Body, TBody).
\end{lstlisting}

We showcase how a reasoning trace is induced using the example below.
Given a knowledge base like:
\begin{lstlisting}[language=Prolog]
parent_of(X, Y) :- mother_of(X, Y).
parent_of(X, Y) :- father_of(X, Y).
grandparent_of(X, Y) :- 
    parent_of(X, Z), parent_of(Z, Y).
mother_of(morty, beth).
father_of(beth, rick).
\end{lstlisting}
and a query:
\begin{lstlisting}[language=Prolog]
?- mi_tree(g(
    grandparent_of(morty, Who)), 
    Proof
    ).
\end{lstlisting}
the output would be
\begin{lstlisting}[language=Prolog]
Who=rick,
Proof=
(
  (
    (
      true
      =>mother_of(morty, beth)
    )
    =>parent_of(morty, beth), 
    (
      true
      =>father_of(beth, rick)
    )
    =>parent_of(beth, rick)
  )
  =>grandparent_of(morty, rick)
).
\end{lstlisting}
The output proof is ensured to be causal and reliable since it is deterministically generated.
A notable advantage is that the reasoning process remains uninfluenced no matter how many distractors are added to the problem description, in contrast to LLMs that can be easily distracted by irrelevant context~\citep{10.5555/3618408.3619699}.

\subsubsection{Search Strategy}
\label{sec:method:symbolic_inference:search_strategy}
The default search strategy of Prolog is DFS, which may lead to infinite loops.
For example, given the knowledge base:
\begin{lstlisting}[language=Prolog]
parent_of(X, Y) :- offspring_of(Y, X).
offspring_of(X, Y) :- parent_of(Y, X).
parent_of(X, Y) :- mother_of(X, Y).
parent_of(X, Y) :- father_of(X, Y).
mother_of(jack, anna).
\end{lstlisting}
and a query \verb|?- parent_of(jack, Who). |, the backtracking process would repeat over the first two lines without resorting to other lines due to DFS.
To address this issue, we adopt Iterative Deepening Search (IDS), in which the backtracking process performs a series of depth-limited searches, each with an increasing depth limit.
This leverages the strengths of both Breadth-First Search (BFS) and DFS.



\section{Experiments}
\label{sec:exp}
We briefly introduce our experimental settings in $\S$\ref{sec:exp:setting} and show the experiment results in $\S$\ref{sec:exp:main_results}.

\subsection{Experimental Settings}
\label{sec:exp:setting}
We present our experimental settings in this section, including our implementation details of the two components ($\S$\ref{sec:exp:setting:implementation}), a brief introduction of the adopted datasets ($\S$\ref{sec:exp:setting:datasets}) and the baselines ($\S$\ref{sec:exp:setting:baselines}).

\subsubsection{Implementation}
\label{sec:exp:setting:implementation}

\paragraph{SymGen}
We adopt the Code-LLaMA~\citep{code-llama} family as the base LLMs to translate natural languages into Prolog representations.
Our prompting paradigm is in a pure few-shot in-context-learning (ICL) prompting style, without detailed human-written instructions.
Each ICL demonstration comprises a question and a piece of Prolog code.

\paragraph{SymInfer}
We adopt SWI-Prolog~\citep{swipl} and PySwip\footnote{\url{https://github.com/yuce/pyswip}} packages to implement the symbolic inference engine.
We set the maximum depth to be 20 for Iterative Deepening Search and the number of generated reasoning paths to 20.

\subsubsection{Datasets}
\label{sec:exp:setting:datasets}

We evaluate \textsc{CaRing} on three popular complex reasoning datasets, including two logical reasoning datasets (ProofWriter~\citep{tafjord-etal-2021-proofwriter} and PrOntoQA~\citep{saparov2023language}) and one arithmetic dataset (GSM8K~\citep{cobbe2021gsm8k,ribeiro2023street}).

\paragraph{ProofWriter}
ProofWriter~\citep{tafjord-etal-2021-proofwriter} is a commonly-used logical reasoning dataset.
It contains many small rulebases of facts and rules, expressed in English. 
Each rulebase has a set of questions (English statements) that can either be proven true or false using proofs of various depths, or the answer is “Unknown” (in open-world setting, OWA).
The proofs can naturally be represented as directed acyclic graphs (DAGs). 
The dataset is divided into several sub-sets according to maximum proof depth, namely \{$0$, $\leq 1$, $\leq 2$, $\leq 3$, $\leq 5$\}.
We follow previous work~\citep{PanLogicLM23} to use a 600-instance subset sampled from the most difficult depth-5 test set.
We also report additional results on the full depth-5 test set in Appendix~$\S$\ref{sec:appendix:add_result:proofwriter}.

\paragraph{PrOntoQA}
PrOntoQA~\citep{saparov2023language} is a synthetic question answering dataset designed for diagnosing the logical reasoning ability of LLMs.
Each example aims to validate the feasibility of a statement given a context.
We report results on two subsets so we can compare \textsc{CaRing} with previous methods.
As for the results reported in Table~\ref{tab:exp:main:prontoqa}, we follow \citet{PanLogicLM23} to adopt the most difficult depth-5 \textit{fictional characters} sub-set, which contains 500 statement-context pairs.
As for the results in Table~\ref{tab:exp:main:prontoqa-rap}, we use the subset adopted by \citet{rap}.
Similar to ProofWriter, the proofs provided by the dataset can be naturally represented as DAGs.

\paragraph{GSM8K}
GSM8K~\citep{cobbe2021gsm8k} is a multi-step arithmetic reasoning dataset composed of high-quality grade school math word problems.
The original GSM8K dataset contains reasoning explanations written in natural language, which raises difficulties in evaluating intermediate steps automatically.
Recently, \citet{ribeiro2023street} released a subset that contains 270 questions annotated with structured reasoning proofs in the format of DAGs.
We adopt this subset to enable the evaluation of reasoning proofs.

\subsubsection{Evaluation Metrics}
\label{sec:exp:implementation:metrics}
\citet{ribeiro2023street} proposed two novel metrics to evaluate the quality of the generated reasoning proofs in addition to the prevalent answer accuracy metric.
Similar to them, we adopt the following metrics to evaluate both the answers and the generated reasoning proofs.

\paragraph{Answer Accuracy}
Answer accuracy measures a model's ability to predict the correct answer.
A prediction is deemed correct if it is (i) the same as the gold option for multi-choice problems and (ii) the same integer as the gold answer for arithmetic reasoning problems.
This metric is the upper bound for other metrics since a reasoning graph would be marked as incorrect without evaluation if the answer is marked as incorrect.
We report this metric for all datasets.



 \begin{table}[t]
    \centering
    \resizebox{0.85\columnwidth}{!}{
    \begin{tabular}{cclccc}
    \specialrule{.12em}{1em}{0em}
         &  & \multirow{2}{*}{\textbf{Method}} & \multirow{2}{*}{\textbf{Acc (\%)}} & \multicolumn{2}{c}{\textbf{Proof Sim (\%)}}\\
        & & & & All & Correct \\
        \hline
        \parbox[t]{2mm}{\multirow{5}{*}{\rotatebox[origin=c]{270}{\textbf{GPT-4*}}}} 
        &  & CoT & 67.41 & -- & --\\
        &  & ToT & 70.33 & -- & --\\
        &  & CR & 71.67 & -- & --\\
        &  & DetermLR & 79.17 & -- & --\\
        &  & Logic-LM & 79.66 & -- & --\\
        \hline
        \parbox[t]{2mm}{\multirow{6}{*}{\rotatebox[origin=c]{270}{\textbf{Code-LLaMA}}}}
        & \parbox[t]{2mm}{\multirow{2}{*}{\rotatebox[origin=c]{270}{\textbf{7B}}}} & CoT & 46.33 & 9.69 & 14.95\\
        & & Ours & 91.00 & 72.91 & 84.39 \\
        \cdashline{3-6}[1.2pt/2pt]
        & \parbox[t]{2mm}{\multirow{2}{*}{\rotatebox[origin=c]{270}{\textbf{13B}}}} & CoT & 46.50 & 15.69 & 25.86 \\
        & & Ours & 95.67 & 80.65 & 86.00 \\
        \cdashline{3-6}[1.2pt/2pt]
        & \parbox[t]{2mm}{\multirow{2}{*}{\rotatebox[origin=c]{270}{\textbf{34B}}}} & CoT & 52.00 & 15.76 & 27.74 \\
        & & Ours & \textbf{96.50} & \textbf{81.02} & \textbf{86.12} \\
    \specialrule{.12em}{.05em}{.0em}
    \end{tabular}
    }
    \caption{
    Results on the subset of ProofWriter adopted by \citet{PanLogicLM23}. The default setting is 2-shot. ``All'': on all instances. ``Correct'': on correctly-predicted instances. *All GPT-4 numbers are from \citet{determlr}.
    }
    \label{tab:exp:main:proofwrter-logiclm}
    \vspace{-6pt}
\end{table}

 \begin{table*}[t!]
    \centering
    \resizebox{0.6\textwidth}{!}{
    \begin{tabular}{lcccccc}
    \specialrule{.12em}{1em}{0em}
         \multirow{2}{*}{\textbf{Base LLM}} & \multirow{2}{*}{\textbf{\#Param}} & \multirow{2}{*}{\textbf{\#Shot}} & \multirow{2}{*}{\textbf{Method}} & \multirow{2}{*}{\textbf{Acc (\%)}} & \multicolumn{2}{c}{\textbf{Proof Acc (\%)}} \\
        & & & & & All & Correct \\
        \hline
        \multirow{2}{*}{{LLaMA-1}*} & \multirow{2}{*}{{33B}} & \multirow{2}{*}{{8-shot}} & CoT & 87.8 & 64.8 & -- \\
        & & & RAP & 94.2 & 78.8 & -- \\
        \cdashline{1-7}[1.2pt/2pt]
        \multirow{2}{*}{{Code-LLaMA}} & \multirow{2}{*}{{13B}} & \multirow{2}{*}{{2-shot}}
        & CoT & 80.2 & 52.4 & 53.4 \\
        & & & Ours & \textbf{99.0} & \textbf{98.2} & \textbf{99.2} \\
    \specialrule{.12em}{.05em}{.0em}
    \end{tabular}
    }
    \caption{
    Results on the PrOntoQA subset that was adopted by RAP~\citep{rap} for comparison with their method.
    The results marked with * are from their paper. 
    }
    \label{tab:exp:main:prontoqa-rap}
\end{table*}

\paragraph{Reasoning Proof Similarity}
As shown in Figure~\ref{fig:method:problem}, the problems that we are interested in naturally compose reasoning proofs in the format of directed acyclic graphs (DAGs).
Reasoning proof similarity $\mathrm{sim}(\mathcal{G}_g, \mathcal{G}_p)$ measures the graph similarity between the gold and the predicted reasoning graphs.
We follow \citet{ribeiro2023street} to adopt the graph edit distance function $\delta (\mathcal{G}_g, \mathcal{G}_p)$.
This function quantifies the graph edit distance by determining the minimum number of operations required over nodes and edges to transform one graph into the other, thereby enabling a comparison of $\mathcal{G}_g$ and $\mathcal{G}_p$ based on their structural similarities.
The reasoning graph similarity is normalized to $[0, 1]$ as:
\begin{equation}
\begin{aligned}
  & \mathrm{sim}(\mathcal{G}_p, \mathcal{G}_g) = \\
  & 1 - \frac{\delta (\mathcal{G}_p, \mathcal{G}_g)}{\max \{|N_p| + |E_p|, |N_g| + |E_g| \}}
\end{aligned}
    \label{eq:method:graph_sim}
\end{equation}
where $|N_p|$ and $|E_p|$ denote the count of nodes and edges, respectively, within the predicted reasoning graph.
A similar notation applies to $|N_g|$ and $|E_g|$, which represent the number of nodes and edges in the gold graph.
Note that the reasoning graph similarity is set to zero if the predicted answer is incorrect.
We report this metric for ProofWriter and GSM8K.

\paragraph{Reasoning Proof Accuracy}
This metric evaluates the exact match between the gold and the predicted reasoning proofs in terms of both reasoning graph structures and textual contents\footnote{Note that our implementation here is simpler than that of \citet{ribeiro2023street} because we only apply this metric to PrOntoQA, which is easy to get evaluated.}.
The reasoning proof accuracy is either 1 or 0 for a single instance, making it a discrete version of reasoning proof similarity.
Since this metric requires the dataset to have structured content to enable automatic evaluation, we can only apply it to \mbox{PrOntoQA}, which is specifically designed for easy parsing of the proofs.

\subsubsection{Baselines}
\label{sec:exp:setting:baselines}

All baselines prompt the LLMs with few-shot in-context-learning (ICL) demonstrations.




\paragraph{Chain-of-Thought (CoT)}
CoT prompting~\citep{cot}  prompts LLMs with ICL demonstrations that contain both intermediate reasoning steps and answers.
It serves as a popular and strong baseline for prompting LLMs to solve problems.

\begin{table}[t!]
    \centering
    \resizebox{0.8\columnwidth}{!}{
    \begin{tabular}{llclcc}
    \specialrule{.12em}{1em}{0em}
        & & \multirow{2}{*}{\textbf{Method}} & \multirow{2}{*}{\textbf{Acc (\%)}} & \multicolumn{2}{c}{\textbf{Proof Acc (\%)}} \\
        & & & {All} & {Correct}  \\
        \hline
        \parbox[t]{2mm}{\multirow{2}{*}{\rotatebox[origin=c]{270}{\textbf{GPT4*}}}} &
        & CoT          & 98.8 & -- & --   \\
        & & Logic-LM         & 83.2 & -- & --  \\
        \hline
        \parbox[t]{2mm}{\multirow{6}{*}{\rotatebox[origin=c]{270}{\textbf{Code-LLaMA}}}} & \parbox[t]{2mm}{\multirow{2}{*}{\rotatebox[origin=c]{270}{\textbf{7B}}}} 
        & CoT          & 52.0 & 24.8 & 28.5   \\
        & & Ours         & 98.8 & 98.4 & 99.6  \\
        \cdashline{2-6}[1.2pt/2pt]
        & \parbox[t]{2mm}{\multirow{2}{*}{\rotatebox[origin=c]{270}{\textbf{13B}}}} 
        & CoT          & 61.0 & 32.2 & 35.9  \\
        & & Ours         & 99.4 & 98.8 & 99.4  \\
        \cdashline{2-6}[1.2pt/2pt]
        & \parbox[t]{2mm}{\multirow{2}{*}{\rotatebox[origin=c]{270}{\textbf{34B}}}} 
        & CoT          & 82.8 & 41.0 & 41.0  \\
        & & Ours         & \textbf{100.0} & \textbf{100.0} & \textbf{100.0}  \\


    \specialrule{.12em}{.05em}{.0em}
    \end{tabular}
    }
    \caption{
    Results on the depth-5 subset of PrOntoQA. The default setting is 2-shot. Results marked with * are GPT-4 results reported by Logic-LM~\citep{PanLogicLM23}.
    }
    \label{tab:exp:main:prontoqa}
    \vspace{-6pt}
\end{table}



        


\begin{table}[t!]
    \centering
    \resizebox{0.7\columnwidth}{!}{
    \begin{tabular}{llccc}
    \specialrule{.12em}{1em}{0em}
        & \multirow{2}{*}{\textbf{Method}} & \multirow{2}{*}{\textbf{Acc}} & \multicolumn{2}{c}{\textbf{Proof Sim (\%)}} \\
        & & & {All} & {Correct} \\
        \hline
        \parbox[t]{2mm}{\multirow{2}{*}{\rotatebox[origin=c]{270}{\textbf{7B}}}} 
        & CoT          & 13.70 & 4.99 & 36.39 \\
        & Ours         & 12.22 & 6.57 & 53.72 \\
        \cdashline{1-5}[1.2pt/2pt]
        \parbox[t]{2mm}{\multirow{2}{*}{\rotatebox[origin=c]{270}{\textbf{13B}}}} 
        & CoT          & 15.56 & 5.76 & 37.03 \\
        & Ours         & 21.48 & 11.66 & \textbf{54.26} \\
        \cdashline{1-5}[1.2pt/2pt]
        \parbox[t]{2mm}{\multirow{2}{*}{\rotatebox[origin=c]{270}{\textbf{34B}}}} 
        & CoT          & 35.19 & 13.04 & 37.07 \\
        & Ours         & \textbf{42.22} & \textbf{22.91} & 54.25 \\

        
    \specialrule{.12em}{.05em}{.0em}
    \end{tabular}
    }
    \caption{Results on GSM8K. The default setting is 5-shot. }
    \label{tab:exp:main:gsm8k}
    \vspace{-6pt}
\end{table}

\paragraph{Logic-LM} Logic-LM~\citep{PanLogicLM23} is a neuro-symblic method that adopts symbolic solvers for logical reasoning problems. The main difference between Logic-LM and \textsc{CaRing} is: Logic-LM adopts various solvers for multiple datasets and only focuses on answer accuracy, while \textsc{CaRing} universally uses one solver (i.e., SWI-Prolog); and more importantly, \textsc{CaRing} showcases how SWI-Prolog interpreters can be customized to generate intermediate reasoning proofs and to adopt various search (i.e., problem-solving) strategies.

\paragraph{Search-based Methods}
We include several search-based methods as baselines.
We directly adopt the released results in their papers, since it is too time-consuming to implement these methods on our own.
For ProofWriter, we compare our method with GPT-4 based Tree-of-Thoughts (ToT;~\citep{yao2023treeofthoughts}), Cumulative Reasoning (CR;~\citep{zhang2023cumulative}), and DetermLR~\citep{determlr}.
We cannot make comparisons on reasoning proofs because these methods only reported reasoning accuracy.
For PrOntoQA, we compare our method with RAP~\citep{rap}, in terms of both reasoning accuracy and reasoning proof accuracy.


\subsection{Main Results}
\label{sec:exp:main_results}
Tables~\ref{tab:exp:main:proofwrter-logiclm}, \ref{tab:exp:main:prontoqa-rap}, \ref{tab:exp:main:prontoqa} and \ref{tab:exp:main:gsm8k} show the experimental results on our adopted datasets.

\paragraph{ProofWriter}
The results on ProofWriter are presented in Table \ref{tab:exp:main:proofwrter-logiclm}.
\textsc{CaRing} demonstrates notable improvements over existing baselines, particularly in terms of reasoning proof similarity.
Utilizing Code-LLaMA-34B, \textsc{CaRing} achieves a remarkable answer accuracy of 96.50\% and a reasoning proof similarity of 81.02\%, significantly surpassing the most powerful method using GPT-4 that obtains an accuracy of 79.66\%.

\paragraph{PrOntoQA}
The results on PrOntoQA are presented in Tables \ref{tab:exp:main:prontoqa-rap} and \ref{tab:exp:main:prontoqa}.
\textsc{CaRing} achieves almost full accuracy with the 13B model.
Comparing with RAP, \textsc{CaRing} obtains better results in terms of both answer accuracy and proof accuracy even using a smaller base LLM and fewer ICL demonstrations.
\textsc{CaRing} also outperforms CoT and Logic-LM that use GPT-4.

\paragraph{GSM8K}
The results on GSM8K are presented in Table \ref{tab:exp:main:gsm8k}.
This dataset is more challenging than the previous two logical reasoning datasets for \textsc{CaRing}, since it is generally believed that symbolic languages are restricted by their limited expressiveness and cannot properly handle the ambiguity in real-world human languages.
Surprisingly, with the 34B model, \textsc{CaRing} outperforms the strong CoT baseline by a large margin and almost doubles the reasoning proof similarity (22.91\% vs. 13.04\%).
We attribute such improvements to increasingly powerful LLMs, which can correctly translate ambiguous human languages into formal symbolic representations.

\subsection{When Reasoning Becomes More Complex}
A key difficulty confronted by reasoning systems is the rapid expansion of possible states as the reasoning process becomes more complex, such as when additional statements are considered or the depth of inference is greater.
To investigate how our method handles more complex reasoning problems, we conduct experiments under two controlled settings:
(1) \textbf{\#Depth $\mathbf{\uparrow}$}: How does the answer accuracy change with \#Depth being $<=0$, $<=1$, $<=2$, $<=3$ and $<=5$, respectively;
(2) \textbf{\#Statements $\mathbf{\uparrow}$}: How does the answer accuracy change with \#Statements being $<=20$ and $> 20$, respectively.

\begin{figure}[t!]
\captionsetup[subfigure]{justification=Centering}
\centering
\begin{subfigure}[t]{\columnwidth}
\centering
    \includegraphics[width=0.8\linewidth]{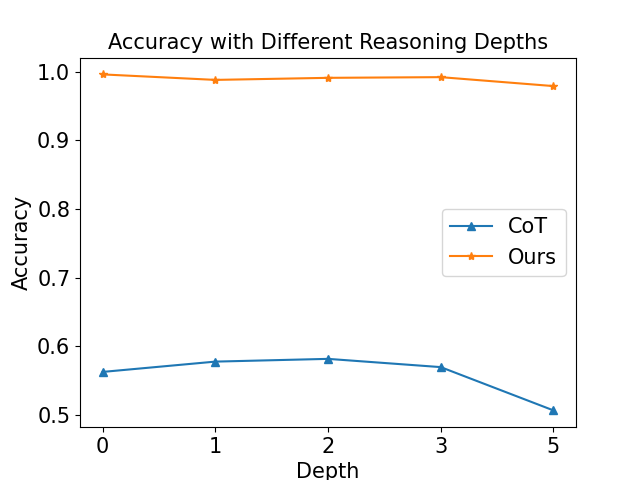}
    \caption{Answer accuracy with different reasoning depths. } \label{subfig:exp:analysis:complex:depth}
\end{subfigure}

\begin{subfigure}[t]{\columnwidth}
\centering
    \includegraphics[width=0.8\linewidth]{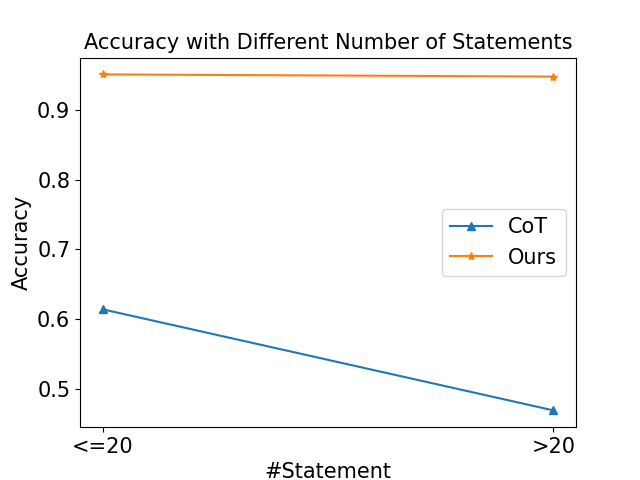}
    \caption{Answer accuracy with different number of statements. } \label{subfig:exp:analysis:complex:statement}
\end{subfigure}

\caption{
        Answer accuracy when reasoning problems become more complex.
        }
\label{fig:exp:analysis:complex}
\end{figure}

As shown in Figures~\ref{subfig:exp:analysis:complex:depth} and \ref{subfig:exp:analysis:complex:statement}, with increasing levels of reasoning intricacy, the answer accuracy of \textsc{CaRing} remains steady.
In contrast, CoT sees significant decreases in answer accuracy under both settings.
This verifies the robustness of \textsc{CaRing} against complex reasoning.

\section{Conclusion}
This paper presents a framework to address the erroneous reasoning proof problem of LLM-based reasoning systems.
Specifically, we develop a neuro-symbolic method called \textsc{CaRing}, which produces high-quality reasoning proofs for complex reasoning problems.
By implementing customized meta-interpreters for executing Prolog representations and putting LLMs under quarantine during the reasoning phase, \textsc{CaRing} ensures the reasoning proofs to be causal and reliable.
We conduct experiments on two logical reasoning datasets and one arithmetic reasoning dataset.
Experimental results demonstrate our method achieves significant improvements with both final answers and intermediate reasoning proofs.
Further analysis indicates \textsc{CaRing} remains robust when the reasoning problems become more complex.

\section*{Limitations}
We observe two limitations regarding our framework:
\begin{itemize}
    \item The generalization ability of \textsc{CaRing} is restricted by the expressiveness of the  concerning symbolic representations. In this paper, we showcase a Prolog-based implementation. Other symbolic representations could be explored to generalize \textsc{CaRing} to more reasoning tasks.
    \item \textsc{CaRing} requires powerful LLMs as symbolic representation generators, which is suggested by the results on GSM8K. This dependence might prevent it from being applied to productions.
\end{itemize}


\bibliography{custom}

\begin{thebibliography}{36}
\expandafter\ifx\csname natexlab\endcsname\relax\def\natexlab#1{#1}\fi

\bibitem[{Andreas et~al.(2016)Andreas, Rohrbach, Darrell, and Klein}]{NeuralModularNetworks}
Jacob Andreas, Marcus Rohrbach, Trevor Darrell, and Dan Klein. 2016.
\newblock \href {https://doi.org/10.1109/CVPR.2016.12} {Neural module networks}.
\newblock In \emph{2016 {IEEE} Conference on Computer Vision and Pattern Recognition, {CVPR} 2016, Las Vegas, NV, USA, June 27-30, 2016}, pages 39--48. {IEEE} Computer Society.

\bibitem[{Besta et~al.(2023)Besta, Blach, Kubicek, Gerstenberger, Gianinazzi, Gajda, Lehmann, Podstawski, Niewiadomski, Nyczyk, and Hoefler}]{graphofthoughts}
Maciej Besta, Nils Blach, Ales Kubicek, Robert Gerstenberger, Lukas Gianinazzi, Joanna Gajda, Tomasz Lehmann, Michal Podstawski, Hubert Niewiadomski, Piotr Nyczyk, and Torsten Hoefler. 2023.
\newblock \href {http://arxiv.org/abs/2308.09687} {Graph of thoughts: Solving elaborate problems with large language models}.

\bibitem[{Cobbe et~al.(2021)Cobbe, Kosaraju, Bavarian, Chen, Jun, Kaiser, Plappert, Tworek, Hilton, Nakano, Hesse, and Schulman}]{cobbe2021gsm8k}
Karl Cobbe, Vineet Kosaraju, Mohammad Bavarian, Mark Chen, Heewoo Jun, Lukasz Kaiser, Matthias Plappert, Jerry Tworek, Jacob Hilton, Reiichiro Nakano, Christopher Hesse, and John Schulman. 2021.
\newblock \href {https://arxiv.org/abs/2110.14168} {Training verifiers to solve math word problems}.
\newblock \emph{ArXiv preprint}, abs/2110.14168.

\bibitem[{Colmerauer and Roussel(1996)}]{prolog}
Alain Colmerauer and Philippe Roussel. 1996.
\newblock \href {https://doi.org/10.1145/234286.1057820} {\emph{The Birth of Prolog}}, page 331–367. Association for Computing Machinery, New York, NY, USA.

\bibitem[{Creswell et~al.(2023)Creswell, Shanahan, and Higgins}]{creswell2023selectioninference}
Antonia Creswell, Murray Shanahan, and Irina Higgins. 2023.
\newblock \href {https://openreview.net/forum?id=3Pf3Wg6o-A4} {Selection-inference: Exploiting large language models for interpretable logical reasoning}.
\newblock In \emph{The Eleventh International Conference on Learning Representations}.

\bibitem[{Dubey et~al.(2024)Dubey, Jauhri, Pandey, Kadian, Al-Dahle, Letman, Mathur, Schelten, Yang, Fan, Goyal, Hartshorn, Yang, Mitra, Sravankumar, Korenev, Hinsvark, Rao, Zhang, Rodriguez, Gregerson, Spataru, Roziere, Biron, Tang, Chern, Caucheteux, Nayak, Bi, Marra, McConnell, Keller, Touret, Wu, Wong, Ferrer, Nikolaidis, Allonsius, Song, Pintz, Livshits, Esiobu, Choudhary, Mahajan, Garcia-Olano, Perino, Hupkes, Lakomkin, AlBadawy, Lobanova, Dinan, Smith, Radenovic, Zhang, Synnaeve, Lee, Anderson, Nail, Mialon, Pang, Cucurell, Nguyen, Korevaar, Xu, Touvron, Zarov, Ibarra, Kloumann, Misra, Evtimov, Copet, Lee, Geffert, Vranes, Park, Mahadeokar, Shah, van~der Linde, Billock, Hong, Lee, Fu, Chi, Huang, Liu, Wang, Yu, Bitton, Spisak, Park, Rocca, Johnstun, Saxe, Jia, Alwala, Upasani, Plawiak, Li, Heafield, Stone, El-Arini, Iyer, Malik, Chiu, Bhalla, Rantala-Yeary, van~der Maaten, Chen, Tan, Jenkins, Martin, Madaan, Malo, Blecher, Landzaat, de~Oliveira, Muzzi, Pasupuleti, Singh, Paluri, Kardas, Oldham, Rita,
  Pavlova, Kambadur, Lewis, Si, Singh, Hassan, Goyal, Torabi, Bashlykov, Bogoychev, Chatterji, Duchenne, Çelebi, Alrassy, Zhang, Li, Vasic, Weng, Bhargava, Dubal, Krishnan, Koura, Xu, He, Dong, Srinivasan, Ganapathy, Calderer, Cabral, Stojnic, Raileanu, Girdhar, Patel, Sauvestre, Polidoro, Sumbaly, Taylor, Silva, Hou, Wang, Hosseini, Chennabasappa, Singh, Bell, Kim, Edunov, Nie, Narang, Raparthy, Shen, Wan, Bhosale, Zhang, Vandenhende, Batra, Whitman, Sootla, Collot, Gururangan, Borodinsky, Herman, Fowler, Sheasha, Georgiou, Scialom, Speckbacher, Mihaylov, Xiao, Karn, Goswami, Gupta, Ramanathan, Kerkez, Gonguet, Do, Vogeti, Petrovic, Chu, Xiong, Fu, Meers, Martinet, Wang, Tan, Xie, Jia, Wang, Goldschlag, Gaur, Babaei, Wen, Song, Zhang, Li, Mao, Coudert, Yan, Chen, Papakipos, Singh, Grattafiori, Jain, Kelsey, Shajnfeld, Gangidi, Victoria, Goldstand, Menon, Sharma, Boesenberg, Vaughan, Baevski, Feinstein, Kallet, Sangani, Yunus, Lupu, Alvarado, Caples, Gu, Ho, Poulton, Ryan, Ramchandani, Franco, Saraf,
  Chowdhury, Gabriel, Bharambe, Eisenman, Yazdan, James, Maurer, Leonhardi, Huang, Loyd, Paola, Paranjape, Liu, Wu, Ni, Hancock, Wasti, Spence, Stojkovic, Gamido, Montalvo, Parker, Burton, Mejia, Wang, Kim, Zhou, Hu, Chu, Cai, Tindal, Feichtenhofer, Civin, Beaty, Kreymer, Li, Wyatt, Adkins, Xu, Testuggine, David, Parikh, Liskovich, Foss, Wang, Le, Holland, Dowling, Jamil, Montgomery, Presani, Hahn, Wood, Brinkman, Arcaute, Dunbar, Smothers, Sun, Kreuk, Tian, Ozgenel, Caggioni, Guzmán, Kanayet, Seide, Florez, Schwarz, Badeer, Swee, Halpern, Thattai, Herman, Sizov, Guangyi, Zhang, Lakshminarayanan, Shojanazeri, Zou, Wang, Zha, Habeeb, Rudolph, Suk, Aspegren, Goldman, Molybog, Tufanov, Veliche, Gat, Weissman, Geboski, Kohli, Asher, Gaya, Marcus, Tang, Chan, Zhen, Reizenstein, Teboul, Zhong, Jin, Yang, Cummings, Carvill, Shepard, McPhie, Torres, Ginsburg, Wang, Wu, U, Saxena, Prasad, Khandelwal, Zand, Matosich, Veeraraghavan, Michelena, Li, Huang, Chawla, Lakhotia, Huang, Chen, Garg, A, Silva, Bell, Zhang, Guo,
  Yu, Moshkovich, Wehrstedt, Khabsa, Avalani, Bhatt, Tsimpoukelli, Mankus, Hasson, Lennie, Reso, Groshev, Naumov, Lathi, Keneally, Seltzer, Valko, Restrepo, Patel, Vyatskov, Samvelyan, Clark, Macey, Wang, Hermoso, Metanat, Rastegari, Bansal, Santhanam, Parks, White, Bawa, Singhal, Egebo, Usunier, Laptev, Dong, Zhang, Cheng, Chernoguz, Hart, Salpekar, Kalinli, Kent, Parekh, Saab, Balaji, Rittner, Bontrager, Roux, Dollar, Zvyagina, Ratanchandani, Yuvraj, Liang, Alao, Rodriguez, Ayub, Murthy, Nayani, Mitra, Li, Hogan, Battey, Wang, Maheswari, Howes, Rinott, Bondu, Datta, Chugh, Hunt, Dhillon, Sidorov, Pan, Verma, Yamamoto, Ramaswamy, Lindsay, Lindsay, Feng, Lin, Zha, Shankar, Zhang, Zhang, Wang, Agarwal, Sajuyigbe, Chintala, Max, Chen, Kehoe, Satterfield, Govindaprasad, Gupta, Cho, Virk, Subramanian, Choudhury, Goldman, Remez, Glaser, Best, Kohler, Robinson, Li, Zhang, Matthews, Chou, Shaked, Vontimitta, Ajayi, Montanez, Mohan, Kumar, Mangla, Ionescu, Poenaru, Mihailescu, Ivanov, Li, Wang, Jiang, Bouaziz,
  Constable, Tang, Wang, Wu, Wang, Xia, Wu, Gao, Chen, Hu, Jia, Qi, Li, Zhang, Zhang, Adi, Nam, Yu, Wang, Hao, Qian, He, Rait, DeVito, Rosnbrick, Wen, Yang, and Zhao}]{dubey2024llama3herdmodels}
Abhimanyu Dubey, Abhinav Jauhri, Abhinav Pandey, Abhishek Kadian, Ahmad Al-Dahle, Aiesha Letman, Akhil Mathur, Alan Schelten, Amy Yang, Angela Fan, Anirudh Goyal, Anthony Hartshorn, Aobo Yang, Archi Mitra, Archie Sravankumar, Artem Korenev, Arthur Hinsvark, Arun Rao, Aston Zhang, Aurelien Rodriguez, Austen Gregerson, Ava Spataru, Baptiste Roziere, Bethany Biron, Binh Tang, Bobbie Chern, Charlotte Caucheteux, Chaya Nayak, Chloe Bi, Chris Marra, Chris McConnell, Christian Keller, Christophe Touret, Chunyang Wu, Corinne Wong, Cristian~Canton Ferrer, Cyrus Nikolaidis, Damien Allonsius, Daniel Song, Danielle Pintz, Danny Livshits, David Esiobu, Dhruv Choudhary, Dhruv Mahajan, Diego Garcia-Olano, Diego Perino, Dieuwke Hupkes, Egor Lakomkin, Ehab AlBadawy, Elina Lobanova, Emily Dinan, Eric~Michael Smith, Filip Radenovic, Frank Zhang, Gabriel Synnaeve, Gabrielle Lee, Georgia~Lewis Anderson, Graeme Nail, Gregoire Mialon, Guan Pang, Guillem Cucurell, Hailey Nguyen, Hannah Korevaar, Hu~Xu, Hugo Touvron, Iliyan Zarov,
  Imanol~Arrieta Ibarra, Isabel Kloumann, Ishan Misra, Ivan Evtimov, Jade Copet, Jaewon Lee, Jan Geffert, Jana Vranes, Jason Park, Jay Mahadeokar, Jeet Shah, Jelmer van~der Linde, Jennifer Billock, Jenny Hong, Jenya Lee, Jeremy Fu, Jianfeng Chi, Jianyu Huang, Jiawen Liu, Jie Wang, Jiecao Yu, Joanna Bitton, Joe Spisak, Jongsoo Park, Joseph Rocca, Joshua Johnstun, Joshua Saxe, Junteng Jia, Kalyan~Vasuden Alwala, Kartikeya Upasani, Kate Plawiak, Ke~Li, Kenneth Heafield, Kevin Stone, Khalid El-Arini, Krithika Iyer, Kshitiz Malik, Kuenley Chiu, Kunal Bhalla, Lauren Rantala-Yeary, Laurens van~der Maaten, Lawrence Chen, Liang Tan, Liz Jenkins, Louis Martin, Lovish Madaan, Lubo Malo, Lukas Blecher, Lukas Landzaat, Luke de~Oliveira, Madeline Muzzi, Mahesh Pasupuleti, Mannat Singh, Manohar Paluri, Marcin Kardas, Mathew Oldham, Mathieu Rita, Maya Pavlova, Melanie Kambadur, Mike Lewis, Min Si, Mitesh~Kumar Singh, Mona Hassan, Naman Goyal, Narjes Torabi, Nikolay Bashlykov, Nikolay Bogoychev, Niladri Chatterji, Olivier
  Duchenne, Onur Çelebi, Patrick Alrassy, Pengchuan Zhang, Pengwei Li, Petar Vasic, Peter Weng, Prajjwal Bhargava, Pratik Dubal, Praveen Krishnan, Punit~Singh Koura, Puxin Xu, Qing He, Qingxiao Dong, Ragavan Srinivasan, Raj Ganapathy, Ramon Calderer, Ricardo~Silveira Cabral, Robert Stojnic, Roberta Raileanu, Rohit Girdhar, Rohit Patel, Romain Sauvestre, Ronnie Polidoro, Roshan Sumbaly, Ross Taylor, Ruan Silva, Rui Hou, Rui Wang, Saghar Hosseini, Sahana Chennabasappa, Sanjay Singh, Sean Bell, Seohyun~Sonia Kim, Sergey Edunov, Shaoliang Nie, Sharan Narang, Sharath Raparthy, Sheng Shen, Shengye Wan, Shruti Bhosale, Shun Zhang, Simon Vandenhende, Soumya Batra, Spencer Whitman, Sten Sootla, Stephane Collot, Suchin Gururangan, Sydney Borodinsky, Tamar Herman, Tara Fowler, Tarek Sheasha, Thomas Georgiou, Thomas Scialom, Tobias Speckbacher, Todor Mihaylov, Tong Xiao, Ujjwal Karn, Vedanuj Goswami, Vibhor Gupta, Vignesh Ramanathan, Viktor Kerkez, Vincent Gonguet, Virginie Do, Vish Vogeti, Vladan Petrovic, Weiwei Chu,
  Wenhan Xiong, Wenyin Fu, Whitney Meers, Xavier Martinet, Xiaodong Wang, Xiaoqing~Ellen Tan, Xinfeng Xie, Xuchao Jia, Xuewei Wang, Yaelle Goldschlag, Yashesh Gaur, Yasmine Babaei, Yi~Wen, Yiwen Song, Yuchen Zhang, Yue Li, Yuning Mao, Zacharie~Delpierre Coudert, Zheng Yan, Zhengxing Chen, Zoe Papakipos, Aaditya Singh, Aaron Grattafiori, Abha Jain, Adam Kelsey, Adam Shajnfeld, Adithya Gangidi, Adolfo Victoria, Ahuva Goldstand, Ajay Menon, Ajay Sharma, Alex Boesenberg, Alex Vaughan, Alexei Baevski, Allie Feinstein, Amanda Kallet, Amit Sangani, Anam Yunus, Andrei Lupu, Andres Alvarado, Andrew Caples, Andrew Gu, Andrew Ho, Andrew Poulton, Andrew Ryan, Ankit Ramchandani, Annie Franco, Aparajita Saraf, Arkabandhu Chowdhury, Ashley Gabriel, Ashwin Bharambe, Assaf Eisenman, Azadeh Yazdan, Beau James, Ben Maurer, Benjamin Leonhardi, Bernie Huang, Beth Loyd, Beto~De Paola, Bhargavi Paranjape, Bing Liu, Bo~Wu, Boyu Ni, Braden Hancock, Bram Wasti, Brandon Spence, Brani Stojkovic, Brian Gamido, Britt Montalvo, Carl
  Parker, Carly Burton, Catalina Mejia, Changhan Wang, Changkyu Kim, Chao Zhou, Chester Hu, Ching-Hsiang Chu, Chris Cai, Chris Tindal, Christoph Feichtenhofer, Damon Civin, Dana Beaty, Daniel Kreymer, Daniel Li, Danny Wyatt, David Adkins, David Xu, Davide Testuggine, Delia David, Devi Parikh, Diana Liskovich, Didem Foss, Dingkang Wang, Duc Le, Dustin Holland, Edward Dowling, Eissa Jamil, Elaine Montgomery, Eleonora Presani, Emily Hahn, Emily Wood, Erik Brinkman, Esteban Arcaute, Evan Dunbar, Evan Smothers, Fei Sun, Felix Kreuk, Feng Tian, Firat Ozgenel, Francesco Caggioni, Francisco Guzmán, Frank Kanayet, Frank Seide, Gabriela~Medina Florez, Gabriella Schwarz, Gada Badeer, Georgia Swee, Gil Halpern, Govind Thattai, Grant Herman, Grigory Sizov, Guangyi, Zhang, Guna Lakshminarayanan, Hamid Shojanazeri, Han Zou, Hannah Wang, Hanwen Zha, Haroun Habeeb, Harrison Rudolph, Helen Suk, Henry Aspegren, Hunter Goldman, Igor Molybog, Igor Tufanov, Irina-Elena Veliche, Itai Gat, Jake Weissman, James Geboski, James Kohli,
  Japhet Asher, Jean-Baptiste Gaya, Jeff Marcus, Jeff Tang, Jennifer Chan, Jenny Zhen, Jeremy Reizenstein, Jeremy Teboul, Jessica Zhong, Jian Jin, Jingyi Yang, Joe Cummings, Jon Carvill, Jon Shepard, Jonathan McPhie, Jonathan Torres, Josh Ginsburg, Junjie Wang, Kai Wu, Kam~Hou U, Karan Saxena, Karthik Prasad, Kartikay Khandelwal, Katayoun Zand, Kathy Matosich, Kaushik Veeraraghavan, Kelly Michelena, Keqian Li, Kun Huang, Kunal Chawla, Kushal Lakhotia, Kyle Huang, Lailin Chen, Lakshya Garg, Lavender A, Leandro Silva, Lee Bell, Lei Zhang, Liangpeng Guo, Licheng Yu, Liron Moshkovich, Luca Wehrstedt, Madian Khabsa, Manav Avalani, Manish Bhatt, Maria Tsimpoukelli, Martynas Mankus, Matan Hasson, Matthew Lennie, Matthias Reso, Maxim Groshev, Maxim Naumov, Maya Lathi, Meghan Keneally, Michael~L. Seltzer, Michal Valko, Michelle Restrepo, Mihir Patel, Mik Vyatskov, Mikayel Samvelyan, Mike Clark, Mike Macey, Mike Wang, Miquel~Jubert Hermoso, Mo~Metanat, Mohammad Rastegari, Munish Bansal, Nandhini Santhanam, Natascha
  Parks, Natasha White, Navyata Bawa, Nayan Singhal, Nick Egebo, Nicolas Usunier, Nikolay~Pavlovich Laptev, Ning Dong, Ning Zhang, Norman Cheng, Oleg Chernoguz, Olivia Hart, Omkar Salpekar, Ozlem Kalinli, Parkin Kent, Parth Parekh, Paul Saab, Pavan Balaji, Pedro Rittner, Philip Bontrager, Pierre Roux, Piotr Dollar, Polina Zvyagina, Prashant Ratanchandani, Pritish Yuvraj, Qian Liang, Rachad Alao, Rachel Rodriguez, Rafi Ayub, Raghotham Murthy, Raghu Nayani, Rahul Mitra, Raymond Li, Rebekkah Hogan, Robin Battey, Rocky Wang, Rohan Maheswari, Russ Howes, Ruty Rinott, Sai~Jayesh Bondu, Samyak Datta, Sara Chugh, Sara Hunt, Sargun Dhillon, Sasha Sidorov, Satadru Pan, Saurabh Verma, Seiji Yamamoto, Sharadh Ramaswamy, Shaun Lindsay, Shaun Lindsay, Sheng Feng, Shenghao Lin, Shengxin~Cindy Zha, Shiva Shankar, Shuqiang Zhang, Shuqiang Zhang, Sinong Wang, Sneha Agarwal, Soji Sajuyigbe, Soumith Chintala, Stephanie Max, Stephen Chen, Steve Kehoe, Steve Satterfield, Sudarshan Govindaprasad, Sumit Gupta, Sungmin Cho, Sunny
  Virk, Suraj Subramanian, Sy~Choudhury, Sydney Goldman, Tal Remez, Tamar Glaser, Tamara Best, Thilo Kohler, Thomas Robinson, Tianhe Li, Tianjun Zhang, Tim Matthews, Timothy Chou, Tzook Shaked, Varun Vontimitta, Victoria Ajayi, Victoria Montanez, Vijai Mohan, Vinay~Satish Kumar, Vishal Mangla, Vlad Ionescu, Vlad Poenaru, Vlad~Tiberiu Mihailescu, Vladimir Ivanov, Wei Li, Wenchen Wang, Wenwen Jiang, Wes Bouaziz, Will Constable, Xiaocheng Tang, Xiaofang Wang, Xiaojian Wu, Xiaolan Wang, Xide Xia, Xilun Wu, Xinbo Gao, Yanjun Chen, Ye~Hu, Ye~Jia, Ye~Qi, Yenda Li, Yilin Zhang, Ying Zhang, Yossi Adi, Youngjin Nam, Yu, Wang, Yuchen Hao, Yundi Qian, Yuzi He, Zach Rait, Zachary DeVito, Zef Rosnbrick, Zhaoduo Wen, Zhenyu Yang, and Zhiwei Zhao. 2024.
\newblock \href {http://arxiv.org/abs/2407.21783} {The llama 3 herd of models}.

\bibitem[{Gupta et~al.(2020)Gupta, Lin, Roth, Singh, and Gardner}]{Gupta2020NeuralModular}
Nitish Gupta, Kevin Lin, Dan Roth, Sameer Singh, and Matt Gardner. 2020.
\newblock \href {https://openreview.net/forum?id=SygWvAVFPr} {Neural module networks for reasoning over text}.
\newblock In \emph{8th International Conference on Learning Representations, {ICLR} 2020, Addis Ababa, Ethiopia, April 26-30, 2020}. OpenReview.net.

\bibitem[{Hao et~al.(2023)Hao, Gu, Ma, Hong, Wang, Wang, and Hu}]{rap}
Shibo Hao, Yi~Gu, Haodi Ma, Joshua Hong, Zhen Wang, Daisy Wang, and Zhiting Hu. 2023.
\newblock \href {https://aclanthology.org/2023.emnlp-main.507} {Reasoning with language model is planning with world model}.
\newblock In \emph{Proceedings of the 2023 Conference on Empirical Methods in Natural Language Processing}, pages 8154--8173, Singapore. Association for Computational Linguistics.

\bibitem[{Huang et~al.(2024)Huang, Chen, Mishra, Zheng, Yu, Song, and Zhou}]{huang2024large}
Jie Huang, Xinyun Chen, Swaroop Mishra, Huaixiu~Steven Zheng, Adams~Wei Yu, Xinying Song, and Denny Zhou. 2024.
\newblock \href {https://openreview.net/forum?id=IkmD3fKBPQ} {Large language models cannot self-correct reasoning yet}.
\newblock In \emph{The Twelfth International Conference on Learning Representations}.

\bibitem[{Hudson and Manning(2019)}]{hudson2019learning}
Drew~A. Hudson and Christopher~D. Manning. 2019.
\newblock \href {https://proceedings.neurips.cc/paper/2019/hash/c20a7ce2a627ba838cfbff082db35197-Abstract.html} {Learning by abstraction: The neural state machine}.
\newblock In \emph{Advances in Neural Information Processing Systems 32: Annual Conference on Neural Information Processing Systems 2019, NeurIPS 2019, December 8-14, 2019, Vancouver, BC, Canada}, pages 5901--5914.

\bibitem[{Kambhampati et~al.(2024)Kambhampati, Valmeekam, Guan, Verma, Stechly, Bhambri, Saldyt, and Murthy}]{kambhampati2024llmscantplanhelp}
Subbarao Kambhampati, Karthik Valmeekam, Lin Guan, Mudit Verma, Kaya Stechly, Siddhant Bhambri, Lucas Saldyt, and Anil Murthy. 2024.
\newblock \href {http://arxiv.org/abs/2402.01817} {Llms can't plan, but can help planning in llm-modulo frameworks}.

\bibitem[{Kazemi et~al.(2023)Kazemi, Kim, Bhatia, Xu, and Ramachandran}]{kazemi-etal-2023-lambada}
Mehran Kazemi, Najoung Kim, Deepti Bhatia, Xin Xu, and Deepak Ramachandran. 2023.
\newblock \href {https://doi.org/10.18653/v1/2023.acl-long.361} {{LAMBADA}: Backward chaining for automated reasoning in natural language}.
\newblock In \emph{Proceedings of the 61st Annual Meeting of the Association for Computational Linguistics (Volume 1: Long Papers)}, pages 6547--6568, Toronto, Canada. Association for Computational Linguistics.

\bibitem[{Kowalski(1979)}]{algorithm_logic_control}
Robert Kowalski. 1979.
\newblock \href {https://doi.org/10.1145/359131.359136} {Algorithm = logic + control}.
\newblock \emph{Commun. ACM}, 22(7):424–436.

\bibitem[{Long(2023)}]{long2023tot}
Jieyi Long. 2023.
\newblock \href {http://arxiv.org/abs/2305.08291} {Large language model guided tree-of-thought}.

\bibitem[{Lyu et~al.(2023)Lyu, Havaldar, Stein, Zhang, Rao, Wong, Apidianaki, and Callison-Burch}]{lyu2023faithful}
Qing Lyu, Shreya Havaldar, Adam Stein, Li~Zhang, Delip Rao, Eric Wong, Marianna Apidianaki, and Chris Callison-Burch. 2023.
\newblock \href {https://arxiv.org/abs/2301.13379} {Faithful chain-of-thought reasoning}.
\newblock \emph{ArXiv preprint}, abs/2301.13379.

\bibitem[{Neelakantan et~al.(2017)Neelakantan, Le, Abadi, McCallum, and Amodei}]{neelakantan2017learning}
Arvind Neelakantan, Quoc~V. Le, Mart{\'{\i}}n Abadi, Andrew McCallum, and Dario Amodei. 2017.
\newblock \href {https://openreview.net/forum?id=ry2YOrcge} {Learning a natural language interface with neural programmer}.
\newblock In \emph{5th International Conference on Learning Representations, {ICLR} 2017, Toulon, France, April 24-26, 2017, Conference Track Proceedings}. OpenReview.net.

\bibitem[{Nye et~al.(2021)Nye, Tessler, Tenenbaum, and Lake}]{nye2021improving}
Maxwell Nye, Michael~Henry Tessler, Joshua~B. Tenenbaum, and Brenden~M. Lake. 2021.
\newblock \href {http://arxiv.org/abs/2107.02794} {Improving coherence and consistency in neural sequence models with dual-system, neuro-symbolic reasoning}.

\bibitem[{OpenAI(2023)}]{gpt4}
OpenAI. 2023.
\newblock \href {http://arxiv.org/abs/2303.08774} {Gpt-4 technical report}.

\bibitem[{Pan et~al.(2023)Pan, Albalak, Wang, and Wang}]{PanLogicLM23}
Liangming Pan, Alon Albalak, Xinyi Wang, and William~Yang Wang. 2023.
\newblock \href {https://arxiv.org/abs/2305.12295} {{Logic-LM:} empowering large language models with symbolic solvers for faithful logical reasoning}.
\newblock In \emph{Findings of the 2023 Conference on Empirical Methods in Natural Language Processing (Findings of EMNLP)}, Singapore.

\bibitem[{Ribeiro et~al.(2023)Ribeiro, Wang, Ma, Zhu, Dong, Kong, Burger, Ramos, zhiheng huang, Wang, Karypis, Xiang, and Roth}]{ribeiro2023street}
Danilo~Neves Ribeiro, Shen Wang, Xiaofei Ma, Henghui Zhu, Rui Dong, Deguang Kong, Juliette Burger, Anjelica Ramos, zhiheng huang, William~Yang Wang, George Karypis, Bing Xiang, and Dan Roth. 2023.
\newblock \href {https://openreview.net/forum?id=1C_kSW1-k0} {{STREET}: A {MULTI}-{TASK} {STRUCTURED} {REASONING} {AND} {EXPLANATION} {BENCHMARK}}.
\newblock In \emph{The Eleventh International Conference on Learning Representations}.

\bibitem[{Rozière et~al.(2023)Rozière, Gehring, Gloeckle, Sootla, Gat, Tan, Adi, Liu, Remez, Rapin, Kozhevnikov, Evtimov, Bitton, Bhatt, Ferrer, Grattafiori, Xiong, Défossez, Copet, Azhar, Touvron, Martin, Usunier, Scialom, and Synnaeve}]{code-llama}
Baptiste Rozière, Jonas Gehring, Fabian Gloeckle, Sten Sootla, Itai Gat, Xiaoqing~Ellen Tan, Yossi Adi, Jingyu Liu, Tal Remez, Jérémy Rapin, Artyom Kozhevnikov, Ivan Evtimov, Joanna Bitton, Manish Bhatt, Cristian~Canton Ferrer, Aaron Grattafiori, Wenhan Xiong, Alexandre Défossez, Jade Copet, Faisal Azhar, Hugo Touvron, Louis Martin, Nicolas Usunier, Thomas Scialom, and Gabriel Synnaeve. 2023.
\newblock \href {http://arxiv.org/abs/2308.12950} {Code llama: Open foundation models for code}.

\bibitem[{Sanyal et~al.(2022)Sanyal, Singh, and Ren}]{sanyal-etal-2022-fairr}
Soumya Sanyal, Harman Singh, and Xiang Ren. 2022.
\newblock \href {https://doi.org/10.18653/v1/2022.acl-long.77} {{F}ai{RR}: Faithful and robust deductive reasoning over natural language}.
\newblock In \emph{Proceedings of the 60th Annual Meeting of the Association for Computational Linguistics (Volume 1: Long Papers)}, pages 1075--1093, Dublin, Ireland. Association for Computational Linguistics.

\bibitem[{Saparov and He(2023)}]{saparov2023language}
Abulhair Saparov and He~He. 2023.
\newblock \href {https://openreview.net/forum?id=qFVVBzXxR2V} {Language models are greedy reasoners: A systematic formal analysis of chain-of-thought}.
\newblock In \emph{The Eleventh International Conference on Learning Representations}.

\bibitem[{Shi et~al.(2023)Shi, Chen, Misra, Scales, Dohan, Chi, Sch\"{a}rli, and Zhou}]{10.5555/3618408.3619699}
Freda Shi, Xinyun Chen, Kanishka Misra, Nathan Scales, David Dohan, Ed~Chi, Nathanael Sch\"{a}rli, and Denny Zhou. 2023.
\newblock Large language models can be easily distracted by irrelevant context.
\newblock In \emph{Proceedings of the 40th International Conference on Machine Learning}, ICML'23. JMLR.org.

\bibitem[{Sun et~al.(2023)Sun, Xu, Liu, Luan, Wang, Shang, Wen, and Yan}]{determlr}
Hongda Sun, Weikai Xu, Wei Liu, Jian Luan, Bin Wang, Shuo Shang, Ji-Rong Wen, and Rui Yan. 2023.
\newblock \href {http://arxiv.org/abs/2310.18659} {From indeterminacy to determinacy: Augmenting logical reasoning capabilities with large language models}.

\bibitem[{Tafjord et~al.(2021)Tafjord, Dalvi, and Clark}]{tafjord-etal-2021-proofwriter}
Oyvind Tafjord, Bhavana Dalvi, and Peter Clark. 2021.
\newblock \href {https://doi.org/10.18653/v1/2021.findings-acl.317} {{P}roof{W}riter: Generating implications, proofs, and abductive statements over natural language}.
\newblock In \emph{Findings of the Association for Computational Linguistics: ACL-IJCNLP 2021}, pages 3621--3634, Online. Association for Computational Linguistics.

\bibitem[{Tafjord et~al.(2022)Tafjord, Dalvi~Mishra, and Clark}]{tafjord-etal-2022-entailer}
Oyvind Tafjord, Bhavana Dalvi~Mishra, and Peter Clark. 2022.
\newblock \href {https://doi.org/10.18653/v1/2022.emnlp-main.134} {Entailer: Answering questions with faithful and truthful chains of reasoning}.
\newblock In \emph{Proceedings of the 2022 Conference on Empirical Methods in Natural Language Processing}, pages 2078--2093, Abu Dhabi, United Arab Emirates. Association for Computational Linguistics.

\bibitem[{Triska(2024)}]{Triska}
Markus Triska. 2024.
\newblock \href {https://www.metalevel.at/acomip/} {A couple of meta-interpreters in prolog}.
\newblock \url{https://www.metalevel.at/acomip/}, Last accessed on 2024-08-16.

\bibitem[{Valmeekam et~al.(2023)Valmeekam, Marquez, Olmo, Sreedharan, and Kambhampati}]{valmeekam2023planbench}
Karthik Valmeekam, Matthew Marquez, Alberto Olmo, Sarath Sreedharan, and Subbarao Kambhampati. 2023.
\newblock \href {https://openreview.net/forum?id=YXogl4uQUO} {Planbench: An extensible benchmark for evaluating large language models on planning and reasoning about change}.
\newblock In \emph{Thirty-seventh Conference on Neural Information Processing Systems Datasets and Benchmarks Track}.

\bibitem[{Wang et~al.(2023)Wang, Wei, Schuurmans, Le, Chi, Narang, Chowdhery, and Zhou}]{selfconsistency}
Xuezhi Wang, Jason Wei, Dale Schuurmans, Quoc~V Le, Ed~H. Chi, Sharan Narang, Aakanksha Chowdhery, and Denny Zhou. 2023.
\newblock \href {https://openreview.net/forum?id=1PL1NIMMrw} {Self-consistency improves chain of thought reasoning in language models}.
\newblock In \emph{The Eleventh International Conference on Learning Representations}.

\bibitem[{Wei et~al.(2022)Wei, Wang, Schuurmans, Bosma, brian ichter, Xia, Chi, Le, and Zhou}]{cot}
Jason Wei, Xuezhi Wang, Dale Schuurmans, Maarten Bosma, brian ichter, Fei Xia, Ed~H. Chi, Quoc~V Le, and Denny Zhou. 2022.
\newblock \href {https://openreview.net/forum?id=_VjQlMeSB_J} {Chain of thought prompting elicits reasoning in large language models}.
\newblock In \emph{Advances in Neural Information Processing Systems}.

\bibitem[{Wielemaker et~al.(2012)Wielemaker, Schrijvers, Triska, and Lager}]{swipl}
Jan Wielemaker, Tom Schrijvers, Markus Triska, and Torbj\"o{}rn Lager. 2012.
\newblock {SWI-Prolog}.
\newblock \emph{Theory and Practice of Logic Programming}, 12(1-2):67--96.

\bibitem[{Yao et~al.(2023)Yao, Yu, Zhao, Shafran, Griffiths, Cao, and Narasimhan}]{yao2023treeofthoughts}
Shunyu Yao, Dian Yu, Jeffrey Zhao, Izhak Shafran, Thomas~L. Griffiths, Yuan Cao, and Karthik Narasimhan. 2023.
\newblock \href {http://arxiv.org/abs/2305.10601} {{Tree of Thoughts}: Deliberate problem solving with large language models}.

\bibitem[{Ye et~al.(2023)Ye, Chen, Dillig, and Durrett}]{ye2023satlm}
Xi~Ye, Qiaochu Chen, Isil Dillig, and Greg Durrett. 2023.
\newblock \href {https://openreview.net/forum?id=TqW5PL1Poi} {Sat{LM}: Satisfiability-aided language models using declarative prompting}.
\newblock In \emph{Thirty-seventh Conference on Neural Information Processing Systems}.

\bibitem[{Ye and Durrett(2022)}]{ye2022the}
Xi~Ye and Greg Durrett. 2022.
\newblock \href {https://openreview.net/forum?id=Bct2f8fRd8S} {The unreliability of explanations in few-shot prompting for textual reasoning}.
\newblock In \emph{Advances in Neural Information Processing Systems}.

\bibitem[{Zhang et~al.(2023)Zhang, Yang, Yuan, and Yao}]{zhang2023cumulative}
Yifan Zhang, Jingqin Yang, Yang Yuan, and Andrew Chi-Chih Yao. 2023.
\newblock \href {https://arxiv.org/abs/2308.04371} {Cumulative reasoning with large language models}.
\newblock \emph{ArXiv preprint}, abs/2308.04371.

\end{thebibliography}

\clearpage

\appendix

\section{Appendix}
\label{sec:appendix}

\subsection{Additional Results}

\begin{table}[t]
    \centering
    \resizebox{1.0\columnwidth}{!}{
    \begin{tabular}{llccc}
    \specialrule{.12em}{1em}{0em}
        & & \multirow{2}{*}{\textbf{Acc}} & \multicolumn{2}{c}{\textbf{Proof Sim}} \\
        & & & \textbf{All} & \textbf{Correct}  \\
        \hline
        \parbox[t]{2mm}{\multirow{5}{*}{\rotatebox[origin=c]{270}{\textbf{7B}}}} 
        & Direct       & 41.78 & -- & --  \\
        & Direct (3-Shot) & 43.32 & -- & --  \\
        & CoT          & 40.95 & 11.27 & 17.20   \\
        & CoT (3-shot) & 42.58 & 11.52 & 21.58   \\
        & Ours         & 92.43 & 75.85 & \textbf{86.68}  \\
        \hline
        \parbox[t]{2mm}{\multirow{5}{*}{\rotatebox[origin=c]{270}{\textbf{13B}}}} 
        & Direct      & 43.44 & -- & --  \\
        & Direct (3-shot) & 44.31 & -- & --   \\
        & CoT          & 45.88 & 16.16 & 27.32  \\
        & CoT (3-shot) & 54.70 & 23.18 & 32.48  \\
        & Ours         & 96.16 & 80.74 & 86.34  \\
        \hline
        \parbox[t]{2mm}{\multirow{5}{*}{\rotatebox[origin=c]{270}{\textbf{34B}}}} 
        & Direct       & 44.00 & -- & --  \\
        & Direct (3-shot) & 45.93 & -- & --  \\
        & CoT          & 52.32 & 15.08 & 26.30  \\
        & CoT (3-shot) & 56.50 & 24.12 & 34.61  \\
        & Ours         & \textbf{98.11} & \textbf{83.17} & 85.65  \\


    \specialrule{.12em}{.05em}{.0em}
    \end{tabular}
    }
    \caption{
    Results on ProofWriter. 
    ``All'' and ``Correct'' refer to ``on all instances'' and ``on correctly-predicted instances'', respectively. ``Proof Sim'' refers to ``Proof Graph Similarity'' while ``Proof EM'' means ``Proof Graph Exact Match''. 
    The default setting is 2-shot. 
    We additionally conduct 3-shot experiments for baselines to include all types of labels in the in-context demonstrations because this dataset contains three labels: \{\texttt{true}, \texttt{false}, \texttt{uncertain}\}. 
    We do not conduct 3-shot experiments for our method because it is not sensitive to the number of labels due to its reasoning-by-execution nature.
    }
    \label{tab:exp:main:proofwriter}
\end{table}

\subsubsection{Results on ProofWriter}
\label{sec:appendix:add_result:proofwriter}
Table~\ref{tab:exp:main:proofwriter} shows the results from our implementation on the depth-5 test set of ProofWriter.

\end{document}